\pdfoutput=1
\documentclass[a4paper]{ceurart}

\usepackage{booktabs}
\usepackage{multirow}
\usepackage{amsmath}
\usepackage{amssymb}
\usepackage{graphicx}
\usepackage{hyperref}
\usepackage{xcolor}
\usepackage{listings}
\usepackage{algorithm}
\usepackage{algpseudocode}
\usepackage{array}
\usepackage{placeins}
\begin{document}

\copyrightyear{2026}
\copyrightclause{Copyright for this paper by its authors.
  Use permitted under Creative Commons License Attribution 4.0
  International (CC BY 4.0).}

\conference{CLEF 2026 Working Notes, 21 -- 24 September 2026, Jena, Germany}

\title{Writerslogic at PAN 2026: Process over Content for Robust Detection under Domain Shift}
\title[mode=sub]{Notebook for the PAN Lab at CLEF 2026}

\author[1]{David Condrey}[%
  orcid=0009-0003-1849-2963,
  email=david@writerslogic.com]
\address[1]{WritersLogic Inc, San Diego, CA, USA}

\begin{abstract}
We describe the Writerslogic systems for three PAN@CLEF 2026 shared tasks---Reasoning Trajectory Detection, Voight-Kampff Generative AI Detection, and Multi-Author Writing Style Analysis---unified by a shared analytical framework: feature robustness under distribution shift is governed by \emph{support overlap} between training and test distributions, not by training-set effect size. This yields a taxonomy (domain-anchored, domain-portable, domain-invariant) that explains why generator-specific features die under domain shift while vocabulary fingerprints (hapax ratio, Yule's K, Heaps' exponent), compression measures, and character n-grams survive. On Reasoning Trajectory Detection, where training was 100\% mathematics and 84\% of test was unseen domains, the framework guided system design to 1st place in source detection (0.85 macro F1 via Opus--Sonnet agreement) and 3rd place in safety classification (0.66 macro F1 via query-refusal decomposition). For Voight-Kampff, we built a calibrated ensemble of DeBERTa-v2 (ONNX), multi-seed LightGBM with 44 domain-portable stylometric features, and SVM on n-gram TF-IDF, combined via learned stacking with isotonic calibration; the best configuration achieved 0.891 on the PAN~2026 test set with exceptionally balanced sub-metrics (0.853--0.902 across all evaluation dimensions), confirming that domain-portable feature design transfers across genres. For Multi-Author Writing Style Analysis, we describe a system fusing spectral clustering over character n-gram similarity graphs, normalized compression distance for local boundary detection, and SmolLM-135M perplexity for neural change-point detection; a platform mix-up meant our run never reached the official evaluation, so we report the design and its a priori predictions rather than measured results. Across all three tasks, features measuring \emph{generation process} properties are designed to outperform features measuring \emph{generated content} properties under domain shift.
Code: \url{https://github.com/dcondrey/voight-kampff-clef2026}, \url{https://github.com/dcondrey/trajectory-detection-clef2026}.
\end{abstract}

\begin{keywords}
AI-generated text detection \sep
domain shift \sep
feature robustness \sep
vocabulary fingerprints \sep
safety classification \sep
ensemble methods
\end{keywords}

\maketitle

\section{Introduction}
\label{sec:intro}

The standard recipe for a text classification shared task is: extract features, train a model, optimize on validation, submit. We followed this recipe on the PAN@CLEF 2026 Reasoning Trajectory Detection task~\cite{bevendorff:2025,bevendorff:2026b}. A LightGBM~\cite{ke2017lightgbm} classifier achieved 0.97 macro F1 on validation but 0.69 on test. An XLM-RoBERTa~\cite{conneau2020xlmr} safety classifier achieved 0.83 on validation but 0.45 on test.

These failures follow from a principle this paper develops: \emph{feature robustness under distribution shift is governed by support overlap---whether the feature has non-trivial values in the test distribution---not by training-set effect size}. This yields a taxonomy (Section~\ref{sec:taxonomy}) and two consequences: the \emph{validation illusion}, where IID validation actively selects non-transferable features; and the \emph{granularity trap}, where fine-grained estimators degrade hierarchical performance until they exceed a computable accuracy threshold.

We participated in three PAN@CLEF 2026 tasks, applying this framework to guide system design:

\begin{enumerate}
  \item \textbf{Reasoning Trajectory Detection} (Section~\ref{sec:trajectory}): Source detection (1st place, 0.85 F1) and safety classification (3rd place, 0.66 F1). The framework was developed retrospectively on this task, where generator-specific features with 0\% test fire rate explained the validation-test gap.
  \item \textbf{Voight-Kampff Generative AI Detection} (Section~\ref{sec:vk}): Cross-genre AI text detection using a calibrated ensemble of DeBERTa-v2, LightGBM, and SVM (0.891 on PAN~2026 test, 0.979 on PAN~2025 test). Feature engineering was informed by the framework: all 44 features were designed to be domain-portable across genres.
  \item \textbf{Multi-Author Writing Style Analysis} (Section~\ref{sec:mawsa}): Paragraph-level style-change detection under adversarial mimicry, fusing spectral clustering over character n-gram graphs, normalized compression distance, and SmolLM-135M perplexity. Our run did not reach the official evaluation platform in time, so we hold no official results; we present the design and register a priori predictions as an untested application of the framework, not a result.
\end{enumerate}

The unifying thesis is that features measuring \emph{generation process} properties are more robust under domain shift than features measuring \emph{generated content} properties.

\section{Related Work}
\label{sec:related}

\paragraph{AI-generated text detection.}
Detection approaches fall into three families. Probability-based methods analyze statistical signatures: DetectGPT~\cite{mitchell2023detectgpt} uses log-probability curvature, GLTR~\cite{gehrmann2019gltr} visualizes token-level rank distributions, and Binoculars~\cite{hans2024binoculars} compares perplexity across models. Stylometric methods extract surface features and train classifiers~\cite{aityan2025lightweight,stamatatos2009survey}. Fine-tuned transformer methods (RoBERTa~\cite{liu2019roberta}, DeBERTa~\cite{he2021deberta}) achieve high in-distribution accuracy but degrade under domain shift~\cite{park2025domain,wang2024m4,li2024mage,xu2023generalization}. Tavan and Najafi~\cite{tavan2024marsan} demonstrated that heterogeneous ensembles outperform individual detectors. Our feature death taxonomy offers one explanation: mixing features from different robustness categories means surviving features compensate when others die.

\paragraph{Cross-domain generalization.}
Park et al.~\cite{park2025domain} propose domain generalization techniques for machine-generated text detection. Liu et al.~\cite{liu2026mgtprism} use spectral alignment for cross-domain transfer. Wang et al.~\cite{wang2026crossdomain} integrate lexical pair dispersion. Mobin and Islam~\cite{mobin2025luxveri} demonstrate inverse perplexity-weighted ensembles. Ramponi and Plank~\cite{ramponi2020domain} observe that decomposition into shift-invariant sub-problems can outperform direct domain adaptation, a principle our safety classification system instantiates.

\paragraph{Vocabulary richness as detection signal.}
Yule's K~\cite{yule1944statistical} measures vocabulary repetitiveness. Heaps' law~\cite{herdan1960type,heaps1978information} captures vocabulary growth rate. Baayen~\cite{baayen2001word} provides theoretical foundations; Tweedie and Baayen~\cite{tweedie1998variable} establish within-author stability across genres. Falt\'{y}nek and Matlach~\cite{faltynek2022hapax} show hapax legomena exhibit within-author regularity. Gambetta et al.~\cite{gambetta2024linguistic} confirm that hapax counts degrade under recursive LLM generation. We extend these to the human/AI distinction across multiple tasks.

\paragraph{LLM safety classification.}
Young~\cite{young2025guardrails} finds top-performing guardrails drop from 91\% to 34\% on novel attacks. Liu et al.~\cite{liu2025calibration} show guardrail models are poorly calibrated. Deng et al.~\cite{deng2024multilingual} demonstrate multilingual queries bypass safety filters. Our 22-language refusal detector addresses the complementary problem of recognizing refusals regardless of language.

\paragraph{Ensemble methods and calibration.}
Stacking~\cite{wolpert1992stacked} trains a meta-learner on base classifier outputs. Isotonic regression~\cite{zadrozny2002transforming} and Platt scaling~\cite{platt1999probabilistic} ensure calibrated probabilities. Wang et al.~\cite{wang2025mixture} propose Mixture-of-Agents for LLM collaboration; our multi-LLM approach uses simpler agreement-based filtering.

\section{Feature Robustness Framework}
\label{sec:taxonomy}

We define \emph{fire rate} as the fraction of test samples where a feature produces a non-default value. Fire rate measures \emph{support overlap}: whether the feature's non-trivial region intersects the test distribution. A feature with 0\% fire rate is \emph{dead}: zero contribution regardless of training-set effect size.

We propose three categories based on our Reasoning Trajectory experiments (Table~\ref{tab:fire-rates}):

\begin{table}[t]
\centering
\caption{Feature fire rates and effect sizes from Reasoning Trajectory Detection. Generator-specific features had strong effect sizes ($d > 1.5$) but 0\% test fire rate; vocabulary fingerprints have moderate effect sizes but 100\% fire rate.}
\label{tab:fire-rates}
\begin{tabular*}{\columnwidth}{@{\extracolsep{\fill}}lcccl@{}}
\toprule
Feature & Train & Test & $d$ & Category \\
\midrule
\texttt{has\_final\_answer}    & 93.6\% & 0.0\%  & 2.12 & Anchored \\
\texttt{reasoning\_tag}        & 99.3\% & 0.0\%  & 1.89 & Anchored \\
\texttt{opens\_with\_the}      & 70.0\% & $\sim$5\% & 1.54 & Anchored \\
\texttt{has\_boxed\_answer}    & 99.0\% & 29.3\% & 0.91 & Portable \\
\texttt{latex\_density}        & 95\%+  & 16\%   & 0.78 & Portable \\
\midrule
\texttt{hapax\_ratio}          & 100\%  & 100\%  & $-$.89 & Invariant \\
\texttt{yules\_k}              & 100\%  & 100\%  & .614 & Invariant \\
\texttt{heaps\_exponent}       & 100\%  & 100\%  & .523 & Invariant \\
\texttt{compression\_ratio}    & 100\%  & 100\%  & .412 & Invariant \\
\texttt{sentence\_length\_cv}  & 98\%   & 97\%   & .750 & Invariant \\
\bottomrule
\end{tabular*}
\end{table}

\paragraph{Domain-Anchored (zero support overlap).}
Features encoding generator- or topic-specific patterns. \texttt{has\_final\_answer} detects DeepSeek-R1's ``\textbf{Final Answer:}'' marker (93.6\% fire rate in training, 0\% in test). These features detect \emph{specific generators writing about specific topics}, not AI-generated text in general.

\paragraph{Domain-Portable (support overlap, degraded separability).}
Features that fire across domains but lose class separation. \texttt{has\_boxed\_answer} detects LaTeX \verb|\boxed{}|, present in 99\% of AI math solutions but only 29\% of non-math test data.

\paragraph{Domain-Invariant (stable support, stable separability).}
Features measuring \emph{generation process} properties. Hapax ratio, Yule's K, and Heaps' exponent differ systematically between human and AI text because LLM next-token sampling produces different tail-frequency distributions than human compositional writing~\cite{zipf1949human,faltynek2022hapax}. Their 100\% fire rate confirms domain independence.

\paragraph{The a priori criterion.}
A feature should be classified as likely anchored if its definition references a specific generator, output format, or topic. This requires no test data, only introspection about the feature's construction. We applied this criterion when designing the Voight-Kampff feature set.

\paragraph{The validation illusion.}
IID validation selects features with high training-distribution support, including anchored features that die under shift. On Reasoning Trajectory Subtask~2, the model with the highest validation F1 (XLM-R, 0.834) produced the lowest test F1 (0.448), a gap of $-$0.386. The diagnostic: when shift is known in advance, validation gains that do not transfer signal feature death, not feature improvement.

\section{Task 1: Reasoning Trajectory Detection}
\label{sec:trajectory}

The PAN@CLEF 2026 Reasoning Trajectory Detection task~\cite{bevendorff:2026b,ta:2026} defines two subtasks. \textbf{Subtask~1 (Source Detection)}: classify a solution as human-written or AI-generated (macro F1). \textbf{Subtask~2 (Safety Classification)}: given a query and AI reasoning trace, produce a trace-level binary label and per-step safety scores.

\subsection{Distribution Shift}

\begin{figure}[t]
\centering
\includegraphics[width=0.82\columnwidth]{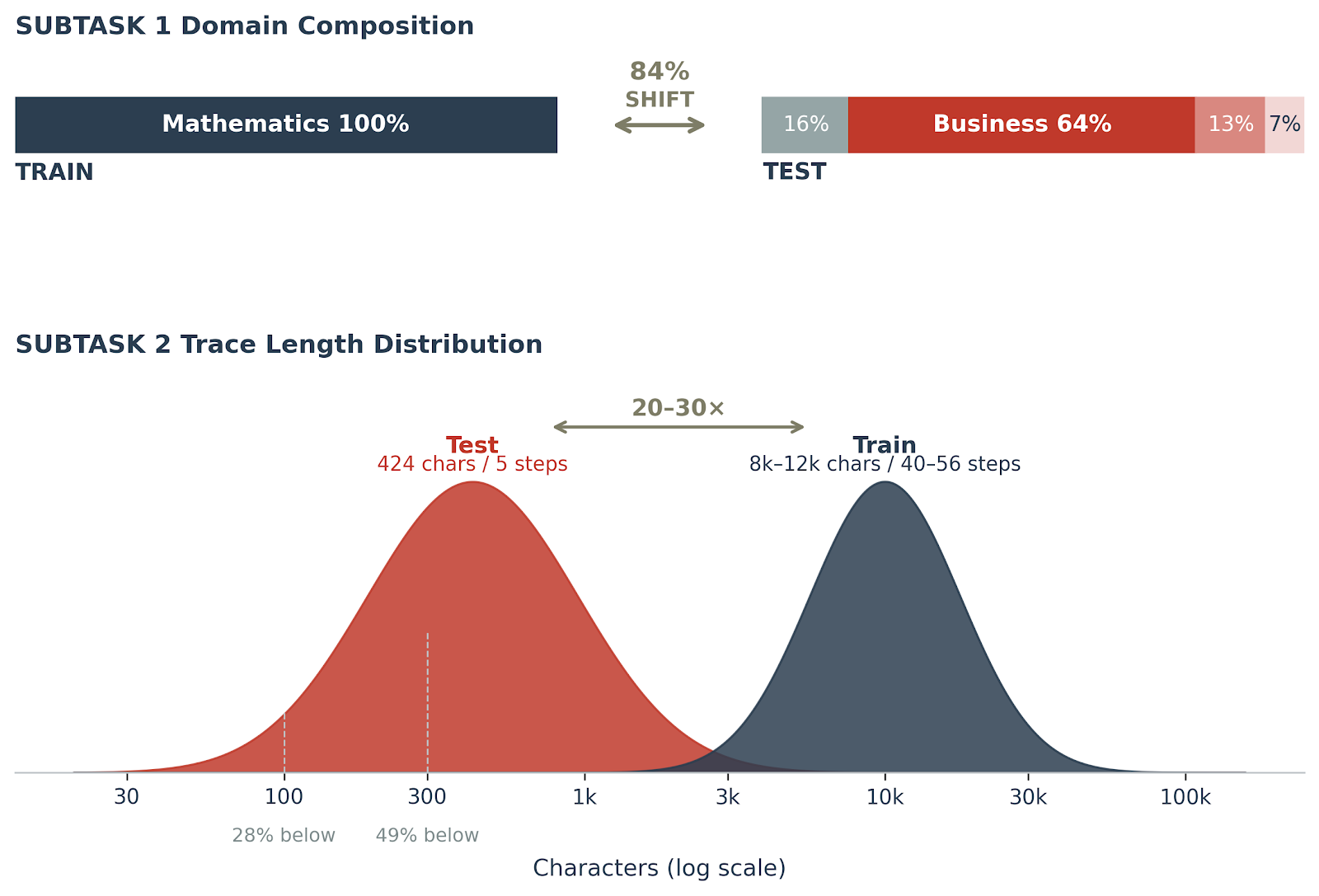}
\caption{Subtask~1 (top): training was 100\% mathematics; 84\% of test is unseen domains. Subtask~2 (bottom): test traces average 424 characters across 15+ languages vs.\ 8,000--12,000 in training.}
\label{fig:cliff}
\end{figure}

Training was 100\% mathematics; 84\% of the test set comprised unseen domains (business 64\%, finance 13\%, coding 7\%). For Subtask~2, training traces averaged 40--56 steps in a single language; test traces averaged 5 steps across 15+ languages and 4 script families (Figure~\ref{fig:cliff}).

\subsection{Subtask 1: Source Detection (1st Place, 0.85 F1)}

\paragraph{Feature-based classifier.}
After removing 5 dead generator-specific features and adding 7 vocabulary fingerprints, our LightGBM ensemble (5 seeds, 30 features) achieved val F1 0.886, test F1 0.690. Adding vocabulary fingerprints improved validation by +0.172 but test by only +0.009, illustrating that even domain-invariant features transfer imperfectly.

\paragraph{Multi-LLM ensemble.}
Llama 3.3 70B and Qwen3-235B classify each solution independently via zero-shot prompts. Test F1: 0.720.

\paragraph{Opus--Sonnet agreement (final system).}
Claude Opus 4.6 and Claude Sonnet 4.6~\cite{anthropic2025claude} classify each sample independently. When both agree, their shared label is used; disagreements are broken by LightGBM. Test F1: \textbf{0.850}. The largest gain (+0.10) came from capability-tier agreement, not feature engineering.

\begin{table}[h]
\centering
\caption{Subtask~1 ablation. The largest gain (+0.10) came from Opus--Sonnet agreement.}
\label{tab:s1-ablation}
\begin{tabular*}{\columnwidth}{@{\extracolsep{\fill}}lcr@{}}
\toprule
System & Test F1 & $\Delta$ \\
\midrule
LightGBM (30 features)       & 0.690 & --- \\
Multi-LLM (Llama + Qwen3)    & 0.720 & +.030 \\
Calibrated combination        & 0.750 & +.030 \\
Opus--Sonnet agreement        & \textbf{0.850} & +.100 \\
\bottomrule
\end{tabular*}
\end{table}

\FloatBarrier
\subsection{Subtask 2: Safety Classification (3rd Place, 0.66 F1)}

All supervised approaches collapsed under distribution shift (Table~\ref{tab:val-vs-test}): XLM-RoBERTa achieved 0.834 val but 0.448 test. This is the \emph{validation illusion}: IID validation selects for anchored features that die under shift.

\begin{table}[t]
\centering
\caption{The validation illusion on Subtask~2. Supervised methods show inverse val/test correlation.}
\label{tab:val-vs-test}
\begin{tabular*}{\columnwidth}{@{\extracolsep{\fill}}lccr@{}}
\toprule
Approach & Val & Test & $\Delta$ \\
\midrule
LightGBM (28 feat.)     & .502 & .362 & $-$.14 \\
Refusal heuristic       & .624 & .350 & $-$.27 \\
TF-IDF + LR             & .793 & .500 & $-$.29 \\
XLM-RoBERTa             & .834 & .448 & $-$.39 \\
\midrule
5-model LLM ensemble    & --   & .551 & -- \\
Query-refusal decomp.   & --   & .640 & -- \\
Multi-signal union      & --   & \textbf{.660} & -- \\
\bottomrule
\end{tabular*}
\end{table}

\paragraph{Two-stage query-refusal decomposition.}
Rather than interpreting ambiguous trace content, we decomposed the problem into two distribution-robust sub-questions: (1)~\emph{Is the query harmful?} (Claude Sonnet via Batch API, aggressive prompt, 58\% classified harmful); (2)~\emph{Does the trace contain a refusal?} (100+ compiled regex patterns across 22 languages in 7 categories: direct refusals, apology-led refusals, policy citations, redirections, indirect refusals, safety meta-reasoning, and model self-identification). Combined rule: safe if refusal detected or query benign; unsafe otherwise. Test F1: 0.640.

\paragraph{Multi-signal union (final system).}
Three Claude Sonnet batches with different prompt strategies combined with query-refusal and LLM ensemble via union (unsafe if any signal votes unsafe). Per-step labels assigned uniformly. Test macro F1: \textbf{0.660}. Soft F1: \textbf{0.270} (2nd place overall). We emphasize that the uniform per-step assignment is \emph{not} unsafe-step localization: it copies the trace-level label onto every step and scores well only because the test traces are largely label-coherent (Section~\ref{sec:discussion}). Its soft-F1 reflects the benchmark's step-label distribution, not any ability of our system to locate the unsafe step.

\paragraph{The granularity trap.}
Post-hoc step-level experiments (GPT-4o adversarial extraction, XLM-RoBERTa fine-tuned on 3,000 synthetic multilingual traces) uniformly degraded performance. Uniform labels outperformed every fine-grained alternative because a step-level estimator must exceed accuracy $1 - \epsilon_{\text{trace}}$ to beat uniform propagation; neither GPT-4o nor XLM-R cleared this threshold. This threshold is a property of the benchmark---specifically its trace-level label coherence---not of our system; we therefore report the uniform-label outcome as a benchmark observation and make no claim to step-localization capability.

\paragraph{Reproducibility and model versions.}
All LLM components used fixed model snapshots with greedy decoding (temperature~$0$). Subtask~1 queried Claude Opus~4.6 (\texttt{claude-opus-4-6}) and Claude Sonnet~4.6 (\texttt{claude-sonnet-4-6}) with zero-shot prompts capped at 10 output tokens, plus the open models Llama~3.3~70B and Qwen3-235B (\texttt{qwen3-235b-a22b}) via hosted inference (OpenRouter/Together.ai). Subtask~2 used Claude Sonnet~4.6 through the Anthropic Message Batches API under three prompt strategies (10 output tokens, except a 200-token rationale variant); GPT-4o and GPT-4o-mini were used only for the post-hoc step-extraction experiments of the granularity-trap analysis. On Opus--Sonnet disagreement the ensemble defers to LightGBM; no separate API-failure fallback was triggered for the batch runs. Experiments ran February--May~2026 on a total compute budget of roughly \$43 (Together.ai) and \$5 (Modal) plus Claude Batch API usage, which we did not meter per experiment. Exact prompts and configurations are released in the linked repositories.

\paragraph{Error analysis.}
The failure modes are consistent across subtasks. On Subtask~1, LightGBM errs on the shifted (business/finance/coding) majority of the test set, where its math-calibrated features fall to their default values; the Opus--Sonnet ensemble instead errs on short, low-signal traces where both models default to ``AI'', which is also where the two agree yet are jointly wrong---so agreement removes the easy errors but not these correlated ones. On Subtask~2, the query--refusal rule misclassifies two categories: traces that comply with a harmful query without emitting an explicit refusal token (predicted safe, actually unsafe), and benign queries phrased as adversarial hypotheticals (predicted unsafe, actually safe). Both are failures of the decomposition's assumption that harm is legible at the query or refusal level rather than in the trace body---the case we deliberately declined to model.

\subsection{Leaderboard}

Table~\ref{tab:leaderboard} places our systems on the official PAN~2026 leaderboards~\cite{ta:2026}.\footnote{Our official runs appear on the PAN leaderboards under two submission identities---\texttt{David Condrey} (Reasoning Trajectory Detection) and \texttt{writerslogic-inc} (Voight-Kampff); both denote this team.} On Subtask~1 our Opus--Sonnet agreement system ranks 1st of nine teams (macro F1 0.85, ahead of the next entry at 0.83). On Subtask~2 we rank 3rd of seven on macro F1 (0.66) but 2nd on soft F1 (0.27), behind only the top entry (0.31): the union system sacrifices a little overall accuracy for better-calibrated step scores. The gap between our 1st-place source detection and mid-table safety classification mirrors the difficulty gap between the two subtasks---source detection rewards a high-precision cross-domain signal (LLM agreement), whereas safety classification under shift resists every approach we tried.

\begin{table}[h]
\centering
\caption{Leaderboard results (top 5 each subtask).}
\label{tab:leaderboard}
\begin{tabular*}{\columnwidth}{@{\extracolsep{\fill}}llcccc@{}}
\toprule
 & Team & F1 & Soft F1 & Rank (F1) & Rank (Soft) \\
\midrule
\multirow{5}{*}{\textbf{S1}} & \textbf{davidcondrey} & \textbf{.85} & -- & \textbf{1} & -- \\
 & srikarkashyap  & .83 & -- & 2 & -- \\
 & skyduan        & .80 & -- & 3 & -- \\
 & affliction     & .75 & -- & 4 & -- \\
 & 23020073       & .72 & -- & 5 & -- \\
\midrule
\multirow{5}{*}{\textbf{S2}} & ioanabrr       & .75 & .31 & 1 & 1 \\
 & srikarkashyap  & .70 & .17 & 2 & 3 \\
 & \textbf{davidcondrey} & \textbf{.66} & \textbf{.27} & \textbf{3} & \textbf{2} \\
 & skyduan        & .65 & .16 & 4 & 4 \\
 & threshdsnmj    & .56 & .09 & 5 & 5 \\
\bottomrule
\end{tabular*}
\end{table}

\section{Task 2: Voight-Kampff Generative AI Detection}
\label{sec:vk}

The PAN@CLEF 2026 Voight-Kampff task~\cite{bevendorff:2026b,bevendorff:2026c} requires classifying text as human-written or AI-generated across multiple genres (essays, news, fiction), outputting a continuous probability score in $[0,1]$. Our feature engineering was directly informed by the Reasoning Trajectory experience: the fire-rate analysis there (Section~\ref{sec:taxonomy}) showed that generator-specific features were \emph{dead} out-of-domain (0\% fire rate, despite $d>1.5$ in training) while vocabulary fingerprints and compression measures retained 100\% fire rate and within-author stability across domains. We therefore carried only that evidence forward: every Voight-Kampff feature was selected to be domain-portable, and no feature keyed to a specific generator or genre was included---trading training-set effect size for support overlap with unseen test genres.

\subsection{Feature Engineering}

We extract 44 domain-portable features organized into eight groups (Table~\ref{tab:vk-features}), deliberately avoiding genre-specific signals.

\begin{table}[t]
\centering
\caption{The 44 domain-portable features for Voight-Kampff, organized by category. All features produce meaningful values regardless of genre.}
\label{tab:vk-features}
\begin{tabular*}{\columnwidth}{@{\extracolsep{\fill}}lr>{\raggedright\arraybackslash}p{0.60\columnwidth}@{}}
\toprule
Category & Count & Representative Features \\
\midrule
Document structure   & 7  & word/char/sentence counts, paragraph count \\
Vocabulary richness  & 8  & type-token ratio, hapax ratio, Yule's K, Heaps' exponent, MATTR~\cite{covington2010cutting}, Zipf coeff. \\
Sentence structure   & 6  & avg.\ sent.\ length, sent.\ length CV, sent.\ start diversity, conjunction start ratio \\
Readability          & 4  & Flesch-Kincaid, Coleman-Liau, avg.\ syllables \\
Compression/entropy  & 4  & zlib ratio, char entropy, repetition ratio \\
Style markers        & 9  & punct ratio, quote density, contraction ratio \\
Discourse            & 4  & transition diversity, connective formality \\
Distributional       & 2  & burstiness, intrinsic dimensionality~\cite{levina2004maximum} \\
Perplexity (GPT-2)   & 4  & log-perplexity, burstiness, rank-1 accuracy, binoculars ratio~\cite{hans2024binoculars} \\
\midrule
\textbf{Total}       & \textbf{44+4} & \\
\bottomrule
\end{tabular*}
\end{table}

The vocabulary fingerprints (hapax ratio, Yule's K, Heaps' exponent, MATTR) are the same features that proved domain-invariant on Reasoning Trajectory (Section~\ref{sec:taxonomy}). They measure properties of the generation process---how vocabulary is sampled and reused---rather than the generated content. The compression features (zlib ratio, character entropy) serve the same role: AI-generated text tends to compress better than human text because it is more predictable.

\subsection{System Architecture}

Three complementary classifiers are combined via learned stacking:

\paragraph{DeBERTa-v2 transformer.}
Fine-tuned DeBERTa-v2~\cite{he2021deberta,he2023debertav3} exported to ONNX~\cite{onnxruntime} with INT8 dynamic quantization for CPU inference. 3 epochs, LR $2 \times 10^{-5}$, AdamW, max 512 tokens.

\paragraph{LightGBM with domain-portable features.}
Multi-seed LightGBM~\cite{ke2017lightgbm} operating on the 44 stylometric features concatenated with truncated SVD projections of character n-gram (3--6) and word n-gram (1--2) TF-IDF~\cite{salton1988tfidf}, plus GPT-2 perplexity features when available.

\paragraph{Calibrated SVM.}
Linear SVM with Platt scaling~\cite{platt1999probabilistic} via scikit-learn's \texttt{CalibratedClassifierCV}~\cite{pedregosa2011scikit}, operating on raw sparse character and word n-gram TF-IDF features.

\paragraph{Ensemble stacking.}
Component probabilities $P_\text{lgb}$, $P_\text{svm}$, $P_\text{deb}$ are combined with interaction features ($\max$, $\min$, std, range) via logistic regression stacker, followed by isotonic regression calibration~\cite{zadrozny2002transforming}. Borderline predictions receive a small bias toward the human class; predictions within a narrow margin of 0.5 abstain.

\subsection{Results}

We submitted 11 software configurations to the TIRA evaluation platform~\cite{froebe:2023b}; the official results are compiled in the task overview~\cite{bevendorff:2026c}. The configurations share the same trained base models and differ only in how those components are composed at inference, along five axes exposed by the runtime configuration: (i) which base classifiers enter the ensemble (LightGBM only, LightGBM\,+\,SVM, or all three with DeBERTa); (ii) the fusion rule (learned logistic stacker over component probabilities and their interaction features, versus a fixed weighted average); (iii) whether isotonic calibration is applied to the fused score; (iv) the size of the multi-seed LightGBM bag; and (v) the abstention margin around the 0.5 decision boundary. Table~\ref{tab:vk-results} summarizes representative configurations. The best system (\texttt{large-rank})---all three classifiers, learned stacker, isotonic calibration---achieved \textbf{0.891} ROC-AUC on the PAN~2026 test set with exceptionally balanced sub-metrics: 0.891 (ROC-AUC), 0.853 (C@1), 0.902 (F1), 0.899 (F0.5u), and 0.887 (mean) across evaluation dimensions~\cite{bevendorff:2026c}, placing 7th by mean score. On the PAN~2025 backward-compatibility test set, the same configuration scored 0.979.

\begin{table}[h]
\centering
\caption{Voight-Kampff results across key configurations. Metric~1 is the primary evaluation score. \texttt{large-rank} achieves the best cross-metric balance; \texttt{brown-velocity} peaks on Metric~1 but collapses on secondary metrics, exhibiting the threshold-collapse pattern predicted by the feature robustness framework.}
\label{tab:vk-results}
\begin{tabular}{lcccccc}
\toprule
Configuration & PAN-26 M1 & M2 & M3 & M4 & M5 & M6 \\
\midrule
\texttt{large-rank}      & \textbf{0.891} & \textbf{0.891} & \textbf{0.853} & \textbf{0.902} & \textbf{0.899} & \textbf{0.887} \\
\texttt{brown-velocity}  & 0.880 & 0.519 & 0.475 & 0.463 & 0.678 & 0.603 \\
\texttt{frozen-author}   & 0.708 & 0.636 & 0.493 & 0.492 & 0.699 & 0.606 \\
\texttt{lively-fixed}    & 0.683 & 0.528 & 0.501 & 0.460 & 0.673 & 0.569 \\
\bottomrule
\end{tabular}
\end{table}

The contrast between \texttt{large-rank} and \texttt{brown-velocity} is instructive: \texttt{brown-velocity} achieves a comparable primary score (0.880) but its secondary metrics collapse (0.519, 0.475, 0.463), indicating optimization for a single decision boundary at the expense of calibration. \texttt{large-rank}'s metric symmetry validates the domain-portable feature design and isotonic calibration pipeline.

\section{Task 3: Multi-Author Writing Style Analysis}
\label{sec:mawsa}

The PAN@CLEF 2026 Multi-Author Writing Style Analysis (MAWSA) task~\cite{bevendorff:2026b} requires detecting style changes (author switches) at paragraph boundaries, with an ``easy'' subset and a ``hard'' adversarial-mimicry subset. Signal selection was guided by the feature robustness framework: under mimicry, surface-level features (vocabulary, sentence length) become unreliable because authors consciously adapt them, making them effectively domain-anchored \emph{relative to the mimicry condition}. We therefore fuse three signals chosen for their resistance to conscious control---the same principle that governs adversarial attacks on other authorship channels, where a verification signal is only as robust as it is hard to manipulate (e.g., timing-forgery against keystroke-based detection~\cite{condrey2026insecurity}). \textbf{We did not complete this task's submission on the official platform within the evaluation window, so no official results exist;} the predictions below are a priori registrations of the framework, not measured outcomes.\footnote{Our style-change pipeline was fully operational, but a mix-up between the CodaBench and TIRA submission endpoints meant the MAWSA run was never submitted to TIRA before the deadline.}

\subsection{Signal 1: Spectral Eigen-Cut}
Each paragraph is represented as a TF-IDF vector over character n-grams (bigrams through 4-grams)~\cite{stamatatos2009survey,salton1988tfidf}, capturing sub-word patterns that are difficult to consciously control. We build an affinity matrix $A_{ij} = \mathbf{x}_i^\top \mathbf{x}_j$ and apply spectral clustering~\cite{vonluxburg2007tutorial} with $k=2$. The partition encodes global topological structure: even when adjacent paragraphs are stylistically similar under mimicry, the spectral embedding can separate them if the overall graph supports a two-cluster normalized cut. A boundary is marked where adjacent paragraphs receive different cluster labels.

\subsection{Signal 2: Normalized Compression Distance}
For each adjacent pair $(p_i, p_{i+1})$ we compute $\mathrm{NCD}(p_i,p_{i+1}) = \frac{C(p_i p_{i+1}) - \min(C(p_i),C(p_{i+1}))}{\max(C(p_i),C(p_{i+1}))}$, where $C(\cdot)$ is the zlib compressed length. NCD approximates the normalized information distance~\cite{li2004similarity,cilibrasi2005clustering}, requires no training data, and is thus domain-invariant by construction; it is most effective when paragraphs exceed roughly 200 characters. We threshold dynamically at $\tau = \mu_{\mathrm{NCD}} + 1.5\,\sigma_{\mathrm{NCD}}$.

\subsection{Signal 3: SmolLM-135M Perplexity}
Each paragraph is passed through SmolLM-135M~\cite{allal2024smollm} to extract token-level surprisal; we derive paragraph-level features (mean, variance, trend) and apply kernel change-point detection with an RBF kernel~\cite{truong2020ruptures}. SmolLM-135M was chosen for practical reasons: at $\sim$270\,MB it fits within the containerized TIRA evaluation constraints.

\subsection{Ensemble Fusion and Predictions}
For documents with $n \geq 5$ paragraphs we take the relaxed disjunction of the spectral and NCD signals, with SmolLM perplexity as a secondary cross-check; for short documents ($n < 5$) we fall back to NCD-only detection with a conservative threshold. Because we obtained no official evaluation for this task, we register two a priori predictions from the framework: (1) character n-gram TF-IDF similarity and NCD retain $>$80\% of their easy-subset detection rate on the hard subset, because sub-word typing patterns and compression profiles are hard to mimic; (2) word-level TF-IDF cosine similarity degrades on the hard subset ($<$30\% boundary detection), because mimicking authors adopt similar vocabulary. These are untested predictions, stated so they remain falsifiable against the official results, not claimed findings.

\section{Cross-Task Analysis}
\label{sec:cross-task}

\subsection{Feature Reuse}

The pattern of feature reuse is consistent with the framework: vocabulary fingerprints and compression features were shared across tasks, while content-anchored features (generator markers, used only on Trajectory until the fire-rate analysis removed them) were task-specific. Reuse alone is not evidence of transfer, though---only Reasoning Trajectory and Voight-Kampff were officially evaluated, so any MAWSA overlap records design intent, not a measured result.

\subsection{Shared Architectural Patterns}

Two architectural patterns recur across tasks:

\paragraph{Ensemble diversity.}
Both systems combine heterogeneous classifiers: DeBERTa + LightGBM + SVM (Voight-Kampff), Opus + Sonnet + LightGBM (Trajectory S1). The diversity principle---different classifiers fail on different inputs---is the practical consequence of mixing features from different robustness categories.

\paragraph{Calibration and fusion.}
Voight-Kampff uses isotonic regression and learned stacking. Trajectory S2 uses union voting. Both prioritize combining signals over optimizing individual components.

\paragraph{Process over content.}
Vocabulary fingerprints, compression features, and character n-grams are process-level features that measure \emph{how} text is produced rather than \emph{what} it says. These features are the common thread connecting the evaluated systems.

\section{Discussion}
\label{sec:discussion}

\paragraph{Decomposition as invariance factorization.}
The query-refusal decomposition on Trajectory S2 succeeded by factoring the label into components with different invariance properties: query harmfulness (invariant to trace length), refusal detection (invariant to query language), and trace interpretation (invariant to neither). The first two are individually shift-tractable; the third was avoided entirely~\cite{arjovsky2019invariant}. This principle---decompose along shift dimensions---may apply to other tasks where holistic classification is brittle.

\paragraph{When agreement beats features.}
On Trajectory S1, two frontier LLMs agreeing on a classification (0.85 F1) outperformed 30 engineered features (0.69 F1) by a wide margin. LLMs \emph{read} text rather than measuring surface statistics; their agreement provides a high-precision cross-domain signal. The calibrated combination (0.75) shows the signals are complementary: LightGBM is confident and correct on extreme cases but unreliable in the middle; LLMs are uniformly decent but occasionally wrong on easy cases.

\paragraph{Implications for shared task design.}
Future tasks with intentional distribution shift should consider providing a shift-representative development set or noting that IID validation may be actively misleading. Our results suggest that task designers who want to test generalization should expect participants' validation rankings to invert on test.

\paragraph{What uniform labels reveal.}
Our naive uniform step labels (all 1.0 or all 0.0) placed 2nd on soft F1, implying the test set's step labels are largely trace-coherent. This is a benchmark property, not a system property.

\subsection{Limitations}

\textbf{Single participant.} Both submissions are from the same researcher using the same analytical lens. The framework was developed retrospectively on Reasoning Trajectory; Voight-Kampff results confirm the framework's predictions (domain-portable features achieved balanced cross-genre performance).

\textbf{API dependence.} Our Trajectory systems depend on proprietary LLM APIs (Claude, GPT-4o), introducing cost, latency, and reproducibility challenges.

\textbf{Ensemble mechanism unknown.} We do not know whether the Opus--Sonnet ensemble succeeds due to capability-tier asymmetry, same-family agreement, or simply Opus being a better classifier.

\textbf{Framework granularity.} The three categories are deliberately coarse. Features exist on a continuum of robustness. Feature-level robustness does not guarantee system-level generalization.

\section{Conclusion}
\label{sec:conclusion}

We presented the Writerslogic systems for three PAN@CLEF 2026 shared tasks, unified by a feature robustness framework: robustness under distribution shift is governed by support overlap, not training-set effect size. This yields a practical diagnostic (compute fire rates on out-of-domain text), a taxonomy (anchored, portable, invariant), and two consequences (the validation illusion, the granularity trap). We caution that the framework was developed retrospectively on Reasoning Trajectory and only partly corroborated on Voight-Kampff; its status is a hypothesis supported by two evaluated tasks, not an established law.

The framework guided system design across tasks: vocabulary fingerprints and compression features for both Voight-Kampff and Reasoning Trajectory; query-refusal decomposition for Trajectory safety classification. The distinction between \emph{what} was generated and \emph{how} it was generated proved a useful axis for feature selection.

For practitioners: before deploying a detection system under expected shift, audit features by the a priori criterion, distrust validation gains that exceed test gains, and propagate parent-level estimates rather than introducing noisy child estimators. These principles guided our systems to 1st place on source detection (0.85 F1) and 2nd on safety soft F1 (0.27).

\section*{Declaration on Generative AI}

During the preparation of this work, the author used Claude Code (Anthropic) to assist with code development, experiment orchestration, and literature search. The author used GPT-4o and GPT-4o-mini (OpenAI) to generate synthetic training data and classify test samples for step-level experiments. After using these tools, the author reviewed and edited the content as needed and takes full responsibility for the publication's content.


\setlength{\bibsep}{2.0pt plus 0.4pt}
\bibliography{references}

\end{document}